\documentclass[letterpaper,10pt,conference]{ieeeconf}

\IEEEoverridecommandlockouts                              
\usepackage{amsmath,amsfonts}
\usepackage{algorithmic}
\usepackage{algorithm}
\usepackage{array}
\usepackage{textcomp}
\usepackage{stfloats}
\usepackage{multirow}
\usepackage{url}
\usepackage{verbatim}
\usepackage{booktabs}
\usepackage{graphicx}   
\usepackage{cite}
\usepackage{color}
\usepackage[alpine,misc]{ifsym}  
\usepackage{caption}
\usepackage{subcaption}
\usepackage[backref=page]{hyperref}
\hypersetup{colorlinks=true, citecolor=blue, linkcolor=red, urlcolor=cyan}

\usepackage{xcolor}
\usepackage{pifont}

\newcommand{\std}[1]{\,{\scriptsize\textpm#1}}

\newcommand{\rev}[1]{%
  \iffalse   
    \begingroup\color{blue}\ignorespaces#1\unskip\endgroup
  \else
    \ignorespaces#1\unskip
  \fi}

\begin{document}
\title{AcousticBEV: Acoustic BEV Object Localization for Intelligent Vehicles}
\author{Ziang Liu\textsuperscript{*} \hspace{2cm} Shiming Wang\textsuperscript{*,\textdagger} \hspace{2cm} Joseph Adeola \hspace{2cm} Julian F.P. Kooij
\thanks{
* Equal contribution. \textdagger~Project lead.
}}



\maketitle
\begin{abstract}
\label{sec:abstract}
Modern autonomous driving and driver assistance systems rely predominantly on cameras, radars and LiDARs 
for environmental perception.
\rev{However, these modalities face different environmental limitations,
including poor illumination for cameras and adverse weather
for LiDAR.}
Acoustic sensing, 
particularly using microphone arrays, 
could provide complementary perception capabilities for autonomous driving, but
has remained largely underexplored. 
In this work,
we design a horizontally arranged microphone array that functions as the `ears' of a vehicle, 
enabling localization of nearby moving vehicles in a bird’s-eye view (BEV) representation that captures both their relative direction and distance. 
Using this array, we collect a real-world driving dataset, \textit{AcousticScenes}, 
which contains 2,117 static and 2,863 dynamic samples with synchronized spatial acoustic signals, LiDAR point clouds and 3D bounding box annotations. 
Building on this dataset, we propose \textit{AcousticBEV}, a learning-based method for acoustic BEV object localization, which estimates the center locations of moving traffic participants in BEV \rev{from the microphone-array signals, conditioned on the ego vehicle’s own speed.}
Experimental results demonstrate that AcousticBEV outperforms both classical signal-processing baselines and other learning-based methods, 
achieving $F1$ scores of \rev{47.7\std{1.2}\%} on `Easy' scenarios and \rev{45.1\std{1.4}\%} on `Difficult' cases.
%
Our findings demonstrate that acoustic signals from a low-cost, vehicle-mounted microphone array can localize nearby moving vehicles in both azimuth and distance.
Code and dataset will be released upon acceptance.
\end{abstract}


\section{Introduction}
\label{sec:intro}

\PARstart{M}{odern} self-driving and Advanced Driver Assistance Systems (ADAS) rely on accurate perception of surrounding traffic participants to ensure anticipation of potentially dangerous situations, safe navigation and decision-making. 
Traditionally, this task is addressed using \rev{conventional}
sensors, including cameras, radars, and LiDARs \cite{caesar2020nuscenes,sun2020wod,oscar2025iv}. 
Despite substantial progress over the past decade, 
\rev{these sensing modalities} exhibit inherent limitations in real-world deployments. 
For instance, 
cameras are highly sensitive to illumination variations and perform poorly in low-light or nighttime conditions, 
while the performance of LiDAR can degrade under adverse weather, such as heavy fog.

\begin{figure}[t]
    \centering
    \includegraphics[width=0.95\columnwidth]{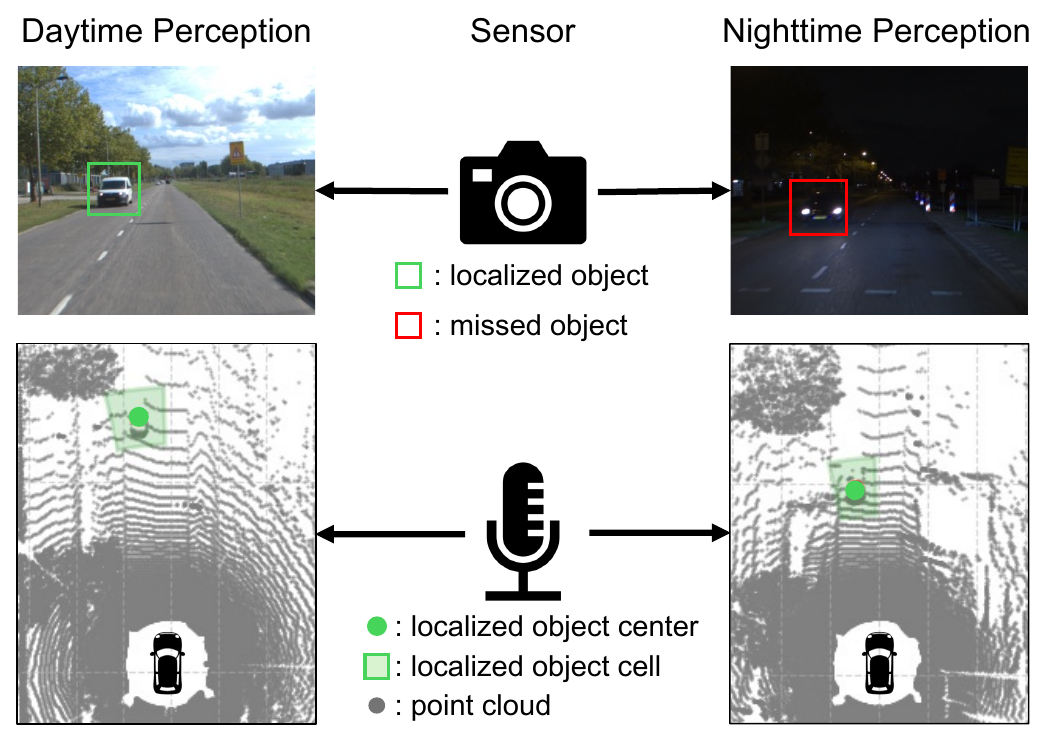}
    \caption{\textbf{Perception comparison between camera and microphone.} 
    \rev{Microphones offer an advantage at night by
capturing acoustic cues independently of illumination,
complementing cameras when visual information is degraded
by low light. The point cloud is shown for reference only.}}
    \label{fig:motivation}
    \vspace{-4mm}
\end{figure}

In contrast,
humans can infer the approximate location of an approaching vehicle solely from acoustic cues, 
even without direct visual observation. 
Inspired by this capability, we ask a fundamental question: 
\textit{Can an intelligent vehicle use microphones (`ears') to detect other traffic participants, even in challenging illumination conditions?}
An illustrative example is shown in Fig.~\ref{fig:motivation}. 
Compared to visual sensors,  microphones can still capture informative signals at night when visual cues from cameras are severely degraded. 

Microphone arrays have already been used in robot perception tasks,
\rev{for example to}
localize speaking people with an indoor humanoid robot \cite{he2018deep},
and to detect sirens for an autonomous vehicle \cite{sun2109emergency}.




The Sound Event Localization and Detection (SELD) task \cite{politis2020overview, mesaros2025decade}
aims to jointly recognize and localize sound events in 3D space from multichannel audio. 
\rev{Although SELD methods such as SELDNet~\cite{adavanne2018seldnet} perform well on benchmarks with stationary indoor arrays, their effectiveness in dynamic driving scenarios remains unclear.}
\rev{In automotive settings, 
prior acoustic work addresses directional or image-plane perception~\cite{gan2019self,chakravarthula2023seeing,valverde2021there,schulz2021hearing}. 
However, the scarcity of 3D data and labels hinders the exploration of 3D perception with acoustic signals.
ARIL~\cite{jeon2024non} was introduced for metric Non-Line-of-Sight (NLoS) vehicle localization, but only evaluated single-target scenarios with a stationary ego vehicle.}

\rev{To address these gaps, we formulate the Acoustic bird's-eye-view (BEV) Object Localization task: estimating the center locations of nearby moving vehicles in BEV space, while the ego vehicle is driving, from a vehicle-mounted microphone array as the sole exteroceptive sensor.}
%
%
To generate data for training and evaluating this task,
we have designed a horizontally arranged two-layer microphone array, consisting of 32 surround microphones for 360$^\circ$ coverage, enabling acoustic sensing in the surrounding BEV space. 
Using this array, we collected a real-world acoustic dataset, \textbf{AcousticScenes}, together with a high-resolution LiDAR sensor. 
Based on the LiDAR measurements, we further annotate 3D bounding boxes for the dynamic objects in the scenes, providing 
accurate object location supervision for learning-based acoustic sensing techniques. 
%
%
\rev{In contrast to existing acoustic datasets~\cite{gan2019self, schulz2021hearing, cantarini2023a3carscene}, 
AcousticScenes targets object-level surround-view perception in real-world driving and is collected under four ego velocities: 0 km/h, 10 km/h, 30 km/h, and 50 km/h.}
%

We further propose \textbf{AcousticBEV}, a novel learning-based method for acoustic BEV object localization.
AcousticBEV integrates the angle and distance uncertainty into the label construction and formulates the task as a polar heatmap prediction problem.
%
\rev{To account for potential variations in acoustic observations
across driving speeds, we introduce an Ego-Velocity Encoding
module that explicitly conditions the network on ego speed.}
This design enables AcousticBEV to become ego-velocity-aware, leading to improved localization performance 
\rev{across various driving speeds.} 

On the AcousticScenes benchmark 
AcousticBEV consistently outperforms both classic and learning-based baselines across various scenarios, 
highlighting the potential of microphone arrays as an effective complementary sensor for intelligent vehicles.

To summarize, this paper's contributions are as follows:
\begin{itemize}
    \item  We present \textbf{AcousticScenes}, a multimodal real-world driving dataset containing synchronized multichannel audio, LiDAR measurements, and 3D bounding box annotations for dynamic traffic participants. 
    \item \rev{\textbf{AcousticBEV} is proposed to localize nearby vehicles in BEV from microphone-array signals, incorporating ego-speed
context through ego-velocity conditioning.}
    \item We comprehensively analyze AcousticBEV in terms of \rev{scene-level analysis}, sensitivity to different ego-vehicle speeds, multi-object localization performance, and the correlation between sound source distance and errors in direction and distance estimation. 
\end{itemize}

\section{Related Work}
\label{sec:related_work}
\textbf{Microphone arrays}
enable the estimation of the direction of incoming sounds. 
\rev{Classic signal processing methods such as beamforming~\cite{sarradj2012three}
and MUSIC~\cite{schmidt1986multiple} estimate Direction-of-Arrival (DOA) from angular
response spectra. Learning-based methods such as
SELDNet~\cite{adavanne2018seldnet} jointly detect sound events
and their directions.} 
\rev{The recurring IEEE DCASE challenge has also included source distance estimation~\cite{dcase2024task3}.}

\textbf{Acoustic perception in robotics}
\rev{provides complementary cues for detecting and localizing
sound-emitting objects~\cite{xiao2015learning}.
Indoor studies often use horizontal microphone
arrays~\cite{he2018deep,kong2020sound,adavanne2018direction,qian2021multi}
to localize alarms~\cite{kong2020sound},
machinery~\cite{cao2019polyphonic}, and
speech~\cite{varzandeh2025improving}.
In human--robot interaction, acoustic cues support speaker
localization~\cite{he2018deep,adavanne2018direction,qian2021multi}.
Beyond source localization, sound cues have been explored for
depth estimation~\cite{brunetto2023audio} and scene geometry
reconstruction~\cite{irie2019seeing,parida2021beyond}.}

\begin{figure}[tb]
    \centering
    \includegraphics[width=0.9\columnwidth]{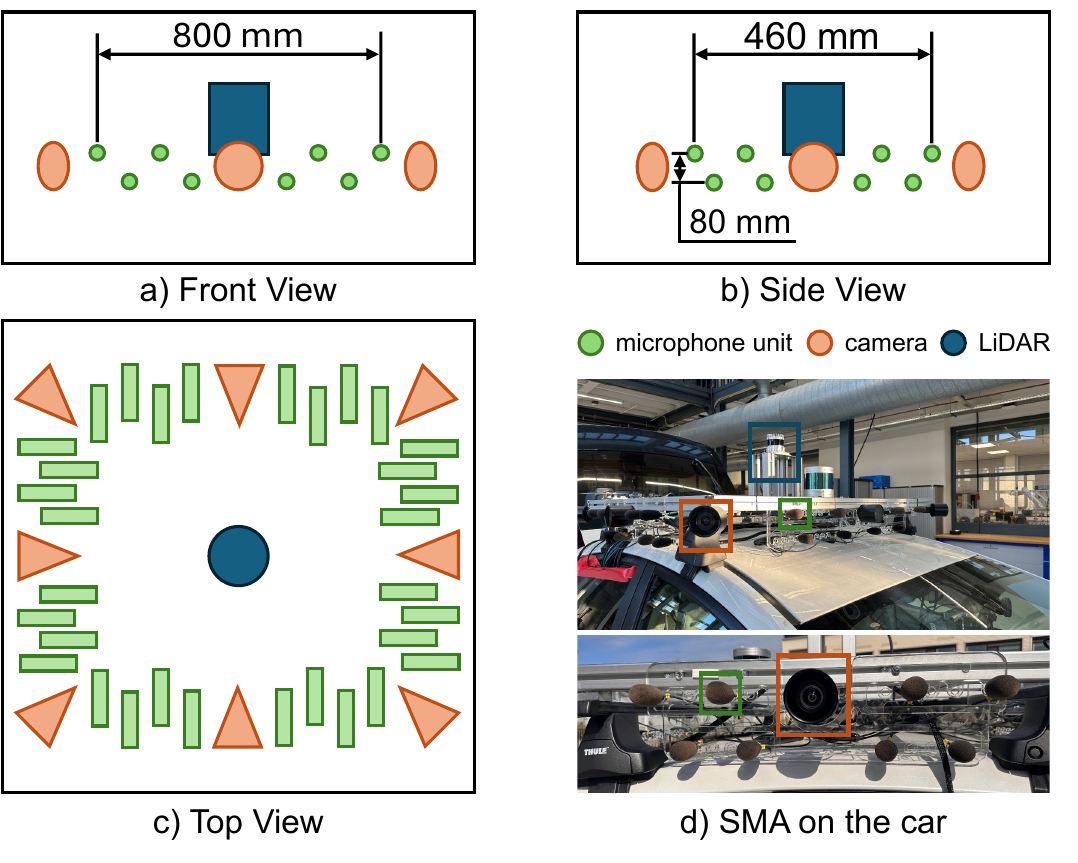}
    \caption{\textbf{Design of the Surround Microphone Array (SMA).} a)-c) Three-view schematic of the layout for 32 microphone units relative to the LiDAR and multi-view cameras. d) Vehicle mounting position.}
    \label{fig:micrphone_design}
    \vspace{-3mm}
\end{figure}

\textbf{Acoustic perception in \rev{intelligent vehicles}} has also gained increasing attention. 
Dai et al. \cite{dai2022binaural} demonstrate that binaural audio can support semantic segmentation of vehicles and obstacles.
\rev{A3CarScene~\cite{cantarini2023a3carscene} provides real-world driving audio for scene-level classification.}
Several works explore vertically arranged 2D microphone arrays
mounted on a vehicle for perception of dynamic traffic scenes~\cite{gan2019self,chakravarthula2023seeing}.
A vertical planar array allows one to estimate the DOA of external sounds in both azimuth and elevation, which can be aligned with a front-facing camera
to complement \rev{image-plane} object detection and tracking.
%
Moreover, 
acoustic perception 
enables awareness beyond the camera’s field of view \cite{valin2003robust}. 
Since sound can reflect and refract around obstacles, various works propose to predict the presence of vehicles in visually occluded regions \cite{schulz2021hearing, hao2024acoustic}. 
\rev{Jeon et al.~\cite{jeon2024non} (ARIL) combine a front-mounted
vertical array with scene geometry for metric NLoS vehicle localization
in BEV, but evaluate only a single vehicle with a stationary ego vehicle
and without 360$^\circ$ surround coverage.}
While datasets with \rev{vertical} arrays primarily focus on perception tasks that align with a front-facing camera view,
$360^\circ$ direction estimation around the vehicle requires horizontal array placement.
\rev{Valverde et al.~\cite{valverde2021there} and Z\"urn~\cite{zurn2022self} use horizontal
microphone arrays to predict image-plane bounding boxes
rather than metric BEV centers.}
\rev{However, the scarcity of datasets pairing acoustic signals with
3D annotations limits the evaluation of acoustic perception
in real-world driving scenarios.}

To bridge this gap, our new dataset contains synchronized acoustic signals 
from a horizontal roof-mounted 32-microphone array
together with high-resolution LiDAR point clouds and accurate 3D annotations of nearby \rev{dynamic} vehicles across diverse driving scenarios.
%
\rev{While various prior acoustic methods focus on image-plane
localization, our proposed AcousticBEV method estimates the angle and distance of one
or more nearby moving vehicles in BEV.}

\section{AcousticScenes dataset}
\label{sec:dataset}

\subsection{Microphone Array Design}

\rev{Prior acoustic perception methods often use large vertical microphone arrays (acoustic cameras), typically calibrated with forward-facing cameras for 2D image-plane detection. Mounted at the vehicle front~\cite{chakravarthula2023seeing} or operated independently~\cite{gan2019self}, these arrays obstruct other onboard sensors and constrain data collection during driving, limiting dataset scalability for realistic onboard deployment.}
%
%
\rev{Motivated by the widespread use of BEV representations in autonomous driving, we design a horizontally arranged, roof-mounted Surround Microphone Array (SMA) to support acoustic BEV perception during driving (Fig.~\ref{fig:micrphone_design}).}

The SMA comprises 32 ADMP441 MEMS microphones mounted on four sides of the vehicle, enabling $360^{\circ}$ spatial awareness.  
Each microphone records acoustic signals at a sampling rate of $48$ kHz with $24$-bit resolution.
On each side, two custom-designed 3D-printed poly(methyl methacrylate) (PMMA) plates are used and each hosts four microphones. 
To improve spatial resolution along the radial and vertical axes, 
we introduce deliberate height and radial offsets between microphones on each side, as illustrated in Fig.~\ref{fig:micrphone_design} a)-c). 
Specifically, one PMMA plate holds four microphones at upper and outer positions, while the other plate holds the remaining four at lower and inner positions.
Each plate features a central groove that enables precise physical alignment with the camera, simplifying cross-sensor calibration for multi-modal perception. 
The SMA is mounted along the roof frame without obstructing the LiDAR field of view or interfering with vehicle operation, 
as shown in Fig.~\ref{fig:micrphone_design} d), 
enabling acoustic data collection in dynamic driving scenarios.

\begin{table}
    \centering
    \caption{Overview of the AcousticScenes Dataset}
    \label{tab:dataset_overview}
    \begin{tabular}
    {    
    >{\centering\arraybackslash}m{1.0cm}
    >{\raggedleft\arraybackslash}m{1.1cm}
    >{\raggedleft\arraybackslash}m{1.45cm}
    >{\raggedleft\arraybackslash}m{1.45cm}
    >{\raggedleft\arraybackslash}m{1.45cm}
    }
    \toprule
    \textbf{Ego Velocity}  & \textbf{\# Scenes} & \textbf{\# Samples} & \textbf{\# Vehicle Samples} & \textbf{\# Silent Samples}  \\
    \midrule
    Static & 65 & 2,117 & 1,530 & 587  \\
    10 km/h & 24 & 615& 476 & 139  \\
    30 km/h & 43 & 1,246 & 990 & 256  \\
    50 km/h & 32 &1,002& 765 & 237  \\
    \midrule
    Total & 164 & 4,980 & 3,761 & 1,219 \\ 
    \bottomrule
    \end{tabular}
\end{table}

\subsection{AcousticScenes Dataset}
%
\rev{As discussed in Section \ref{sec:related_work}, the scarcity of multimodal datasets pairing audio with LiDAR-derived 3D bounding box annotations limits research on acoustic BEV perception in real-world driving scenarios~\cite{irie2019seeing,schulz2021hearing,valverde2021there}.}
To fill this gap, we propose a multimodal acoustic perception dataset, \textbf{AcousticScenes}, 
recorded in Delft, the Netherlands, \rev{containing} synchronized audio data from the SMA and point clouds from a 128-beam rotating LiDAR, Ouster OS1-Rev 7.
%
A \textit{sample} at timestamp $t$ contains a $100$ ms audio clip from $t-100$ ms to $t$, a LiDAR scan at $t$ and the ego-velocity at $t$.
Samples are captured at $5$ Hz. 
Furthermore, we define a \textit{scene} as a sequence of 25--45 consecutive samples, 
corresponding to approximately $5$--$9$ seconds. Each scene represents a complete driving scenario.
\rev{To align with real-world driving, data are recorded both with a
stationary ego vehicle (0 km/h), termed \emph{static scenes}, and while
the ego vehicle drives at 10, 30, and 50 km/h, termed \emph{dynamic scenes}.}
\rev{3D bounding boxes of dynamic objects (`car', `other vehicle', `motorcycle')
are annotated with Segments.ai~\cite{segmentsai2022}.
Category labels are
unused, as this task is class-agnostic.}
%

As summarized in Tab. \ref{tab:dataset_overview}, AcousticScenes comprises $164$ scenes and $4{,}980$ samples in total, 
including $65$ static scenes with $2{,}117$ samples and $99$ dynamic scenes with $2{,}863$ samples. 
%
\rev{We include both \textit{vehicle samples} and 
\textit{silent samples} with no annotated vehicles within the $[-30 \mathrm{m}, +30 \mathrm{m}]$ range. 
Here, `silent' denotes the absence of nearby road users,
rather than the absence of acoustic background noise.}

\textbf{Data Splits.}
The scenes in AcousticScenes have been
split into a \textit{train} set of $130$ ($79\%$) scenes with $3{,}978$ samples, a \textit{val} set of $10$ ($6\%$) scenes with $270$ samples, and a \textit{test} set with $24$  ($15\%$) scenes of $732$ samples.
%
\rev{The \textit{test} scenes are further divided by the number of annotated vehicles: \emph{Easy} scenes contain 0 or 1, and \emph{Difficult} scenes contain 2 or 3.}
%

\begin{figure*}[t]
    \centering
    \includegraphics[width=0.9\textwidth]{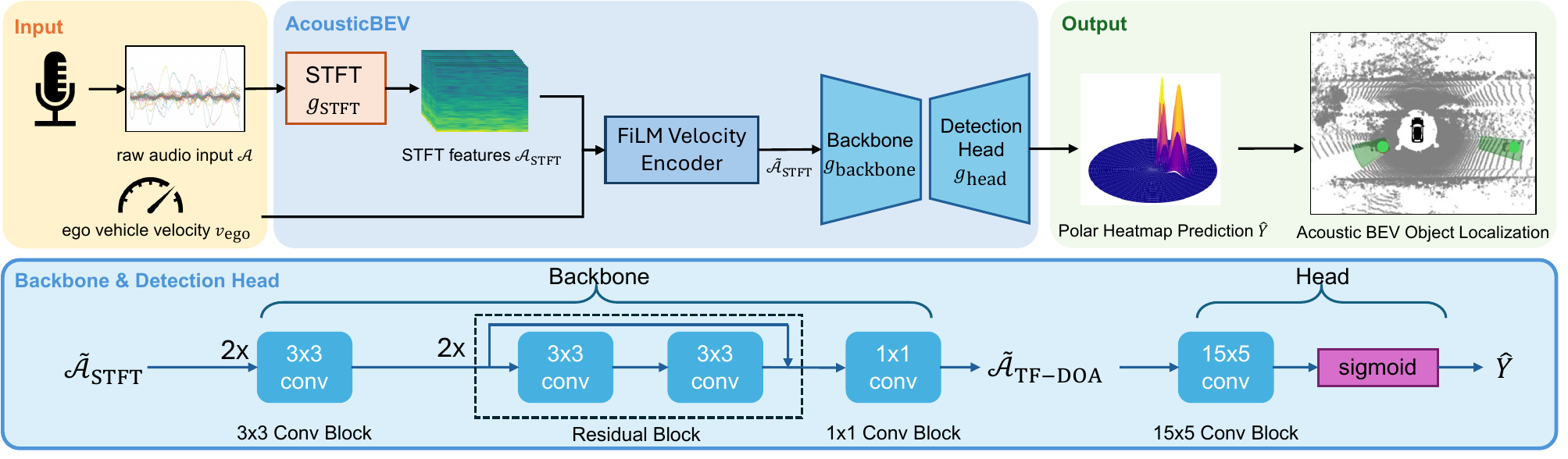}
    \caption{\textbf{Overview of AcousticBEV.} 
    Multi-channel microphone signals are transformed into STFT features and fused with ego-vehicle velocity embeddings. 
    The fused features are fed into a network consisting of a backbone and a detection head, which predicts a polar heatmap. 
    The predicted heatmap is then decoded into object centers.}
    \label{fig:proposed_framework}
\end{figure*}

\section{Methodology}
\label{sec:method}
In this section, we first formalize the Acoustic BEV Object Localization task in Sec. \ref{subsec:task}, 
and then present our method AcousticBEV in Sec. \ref{subsec:acousticbev}.  

\subsection{Acoustic BEV Object Localization}
\label{subsec:task}
\rev{We investigate the potential of acoustic signals for spatial perception in autonomous driving through object-center localization.}
\rev{Compared with camera images and LiDAR point clouds, acoustic signals generally provide less detailed spatial and object-type cues: 
they support direction
and position estimation for moving objects, but offer limited information
about object categories and fine-grained attributes.}
\rev{Therefore,}
we define \textbf{Acoustic BEV Object Localization} as a simplified variant of 3D object detection. 
Instead of regressing full 3D bounding boxes, 
this task predicts \rev{class-agnostic} object centers in BEV \rev{from the multichannel acoustic signal as the only exteroceptive sensor}.

Let $\mathcal{A} \in \mathbb{R}^{M \times (f \cdot \Delta t)}$ denote the synchronized multichannel acoustic signal, where $M$ is the number of microphones, $\Delta t$ denotes a temporal sampling window and $f$ is the inherent sampling rate of the microphones.
Since acoustic perception is naturally associated with direction and distance estimation, we represent each object location in a polar coordinate system.
Formally, let
$
\mathcal{C} = \left\{ \mathbf{p}_n = (\theta_n, r_n) \in \mathbb{R}^{2} \mid n = 1, \dots, N \right\}
$
denote the ground-truth set of object center locations, where $\theta_n$ and $r_n$ are the azimuth angle and distance of the $n$-th object relative to the ego vehicle, and $N \geq 0$ is the number of objects (note that $\mathcal{C} = \emptyset$ when $N = 0$). 
The task is to learn a mapping
$\mathcal{F}: \mathcal{A} \rightarrow \hat{\mathcal{C}}$,
where
$
\hat{\mathcal{C}} = \left\{ \hat{\mathbf{p}}_k = (\hat{\theta}_k, \hat{r}_k) \in \mathbb{R}^{2} \mid k = 1, \dots, K \right\}
$
is the predicted set of object center locations, and $K \geq 0$ is the number of predicted objects.

\subsection{AcousticBEV}
\label{subsec:acousticbev}
\rev{This section describes the proposed AcousticBEV method to address} the \emph{Acoustic BEV Object Localization} task. 
\rev{First, the task is formulated as a dense polar heatmap prediction problem with an uncertainty-aware ground-truth field based on 2D polar Gaussians.} 
\rev{Then, the network architecture and the velocity encoding module are described,}
\rev{followed by the loss function and the object decoding procedure.}

\textbf{Approach: Dense Polar Heatmap Prediction.}
Following the existing center-based detection paradigms~\cite{zhou2019centernet, Duan_2019_ICCV},
\rev{the acoustic BEV object localization task is treated} as a dense prediction task over a discretized polar space.
We define the perception field \rev{in polar space} as $\Theta=[\Theta_{\min}, \Theta_{\max}]$ and $R = [R_{\min}, R_{\max}]$. 
%
\rev{This polar BEV space is discretized into
$N_\theta \times N_r$ cells, with centers
$\theta_i=\Theta_{\min}+(i-\frac{1}{2})\Delta_\theta$,
$i=1,\ldots,N_\theta$, and
$r_j=R_{\min}+(j-\frac{1}{2})\Delta_r$,
$j=1,\ldots,N_r$, where
$\Delta_\theta=(\Theta_{\max}-\Theta_{\min})/N_\theta$
and $\Delta_r=(R_{\max}-R_{\min})/N_r$.}
In \rev{acoustic BEV object localization}, 
%
\rev{given the multichannel acoustic input $\mathcal{A}$, a network $\mathcal{F}_{\phi}$ predicts a dense polar
heatmap
$\hat{Y}=\mathcal{F}_{\phi}(\mathcal{A})
\in[0,1]^{N_\theta\times N_r}$, where $\hat{Y}_{ij}$ denotes
the confidence that an object center lies in cell $(i,j)$.}

\textbf{Label Construction: 2D Polar Gaussian.}
%
\rev{Direct supervision with discrete object centers can lead to unstable training. 
This challenge is compounded by range-dependent uncertainty in acoustic localization.}
\rev{Hence, inspired by CenterPoint~\cite{yin2021centerpoint}, we represent each object center with a 2D Gaussian in polar coordinates, termed a \textit{2D polar Gaussian}, to construct a smooth ground-truth heatmap that accommodates uncertainty in angle and distance estimation.}


\rev{
Let $\mathbf{p}_n=(\theta_n,r_n)$ denote the polar coordinates of the
$n$-th ground-truth object center. We construct its soft label using
a \textit{2D polar Gaussian} kernel centered at $\mathbf{p}_n$:
\begin{equation}
G_n(\theta,r)=\exp\!\left(
-\frac{(\theta-\theta_n)^2}{2\sigma_{\theta_n}^2}
-\frac{(r-r_n)^2}{2\sigma_{r_n}^2}
\right),
\label{eq:2d_gaussian_label}
\end{equation}
where $(\theta,r)$ denotes a query location on the polar BEV grid,
i.e., the center ($\theta_i$, $r_j$) of cell $(i, j)$,
and $\sigma_{\theta_n}$ and $\sigma_{r_n}$ encode uncertainty in
angle and distance estimation, respectively.
}

\rev{
To encode range-dependent localization uncertainty, we model the
angular and radial standard deviations as decreasing and increasing
linearly with the distance $r_n$, respectively. This
parameterization is motivated by the larger angular extent of nearby
objects and greater distance ambiguity at longer ranges:
\begin{equation}
\label{eq:signma_r}
    \sigma_{\theta_n} = \sigma_\theta^{\max} - (r_n/R_{\max})\,(\sigma_\theta^{\max}-\sigma_\theta^{\min}),
\end{equation}
\begin{equation}
\label{eq:signma_theta}
\sigma_{r_n}
=\sigma_r^{\min}
+(r_n/R_{\max})\,
\left(\sigma_r^{\max}-\sigma_r^{\min}\right),
\end{equation}

where $R_{\max}$ denotes the maximum BEV range, and the superscripts
$\min$ and $\max$ denote the positive lower and upper bounds of each
standard deviation.
}

$\sigma_\theta^{\max}$, $\sigma_\theta^{\min}$, $\sigma_r^{\min}$, and $\sigma_r^{\max}$ are hyperparameters and remain constant across all objects and samples. 
\rev{Eqs.~\eqref{eq:signma_r} and \eqref{eq:signma_theta} parameterize
distance-dependent angular and radial uncertainty linearly. 
In practice, localization uncertainty is also influenced
by array aperture, signal-to-noise ratio, and acoustic propagation
conditions, and its dependence on distance may be nonlinear.
The proposed linear formulation offers a simple and interpretable
way to encode the assumed uncertainty trends in the ground-truth
heatmaps.}


\begin{figure}[!b]
    \centering
    \includegraphics[width=0.8\columnwidth]{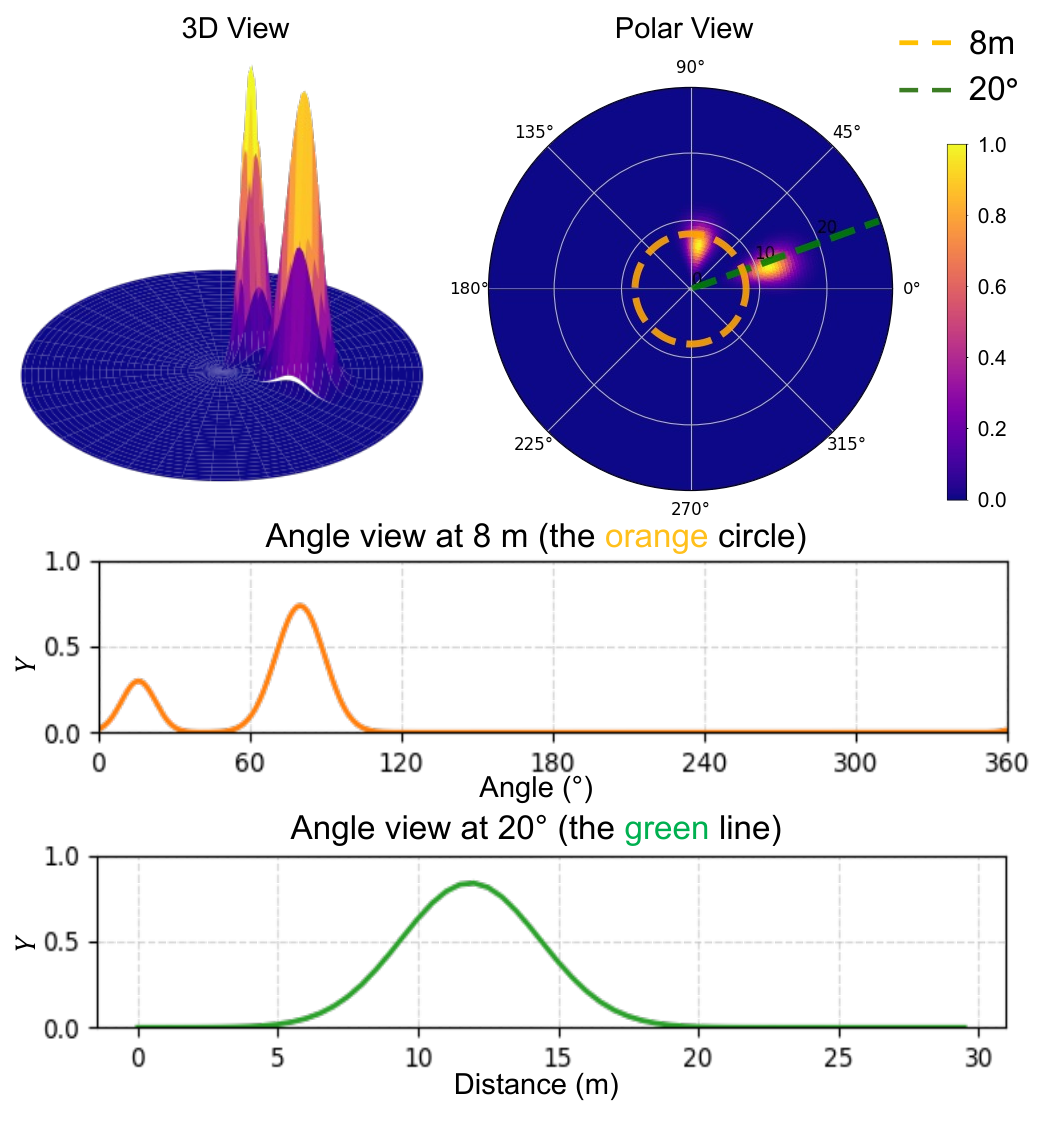}
    \caption{\textbf{2D Polar Gaussian representation.}}
    \label{fig:2d_gaussian}
\end{figure}

\rev{
For a sample with ground-truth object centers
$\mathcal{C}=\{(\theta_n,r_n)\}_{n=1}^{N}$, 
the ground-truth heatmap is constructed by taking
the pointwise maximum over these kernels:
\begin{equation}
    Y_{ij} = \max_{n=1,\dots,N} G_n(\theta_i,r_j).
\end{equation}
Fig.~\ref{fig:2d_gaussian} illustrates the resulting heatmap
for a sample containing two objects.
}

\rev{The proposed network $\mathcal{F}_\phi$ is referred to as
\textbf{AcousticBEV} and its architecture is described below.}

\begin{table*}[!ht]
\caption{\textbf{Acoustic BEV Object-Center Localization on AcousticScenes \textit{test} split.}
$F1^{5^\circ}$  and mAAE of BF and MUSIC are only calculated for angle estimation within the threshold $5^\circ$ and are
excluded from the joint-localization ranking, 
while $F1^{5^\circ,5\mathrm{m}}$, mADE, and mAAE are calculated for acoustic BEV object localization (angle and distance) within thresholds $5^\circ$ and $5 \mathrm{m}$.
\rev{Learning-based methods: mean \textpm\ std over three training seeds.}
(\textbf{Best}, \underline{Second Best}).
}
\label{tab:compare_baseline}
\centering
\resizebox{\textwidth}{!}{
\begin{tabular}{l|c|c|ccc|c|ccc}
\toprule
\multirow[b]{2}{*}{\textbf{Method}}
& \multirow[b]{2}{*}{\textbf{Params (M)}}
& \multicolumn{4}{c|}{\textbf{Easy} \rev{(13 scenes, 322 samples)}}
& \multicolumn{4}{c}{\textbf{Difficult} \rev{(11 scenes, 410 samples)}} \\
\cmidrule(lr){3-6}\cmidrule(lr){7-10}
& & $F1^{5^\circ}$ (\%) $\uparrow$
& $F1^{5^\circ,5\mathrm{m}}$ (\%) $\uparrow$
& mADE (m) $\downarrow$
& mAAE ($^\circ$) $\downarrow$
& $F1^{5^\circ}$ (\%) $\uparrow$
& $F1^{5^\circ,5\mathrm{m}}$ (\%) $\uparrow$
& mADE (m) $\downarrow$
& mAAE ($^\circ$) $\downarrow$ \\
\midrule
BF
& --
& \underline{53.8} & --   & --   & 1.95
& 37.3 & --   & --   & 2.15 \\
MUSIC~\cite{schmidt1986multiple}
& --
& 45.2 & --   & --   & 1.76
& \underline{50.5} & --   & --   & 2.06 \\
\midrule
M-BF+Head
& 0.6
& \rev{41.5\std{1.2}} & \rev{30.1\std{1.1}} & \rev{2.35\std{0.12}} & \rev{\underline{2.14}\std{0.38}}
& \rev{28.5\std{3.9}} & \rev{21.6\std{3.2}} & \rev{2.34\std{0.07}} & \rev{\underline{1.93}\std{0.20}} \\
SELDNet~\cite{adavanne2018seldnet,shimada2022multi}
& 0.8
& \rev{51.2\std{4.3}}
& \rev{\underline{33.1}\std{5.7}}
& \rev{\underline{2.14}\std{0.15}}
& \rev{2.33\std{0.28}}
& \rev{36.5\std{2.2}}
& \rev{\underline{25.5}\std{2.6}}
& \rev{\underline{2.25}\std{0.32}}
& \rev{2.34\std{0.16}} \\
\textbf{AcousticBEV (ours)}
& 3.3
& \rev{\textbf{56.3}\std{1.6}} & \rev{\textbf{47.7}\std{1.2}} & \rev{\textbf{1.76}\std{0.13}} & \rev{\textbf{1.79}\std{0.09}}
& \rev{\textbf{52.1}\std{1.5}} & \rev{\textbf{45.1}\std{1.4}} & \rev{\textbf{1.81}\std{0.06}} & \rev{\textbf{1.85}\std{0.05}} \\
\bottomrule
\end{tabular}
}
\end{table*}

\textbf{Feature Extraction.}
Following existing work \cite{he2018deep}, 
we apply the short-time Fourier transform (STFT) $g_{\text{STFT}}(\cdot)$ to convert the raw acoustic signal into a time–frequency representation, 
which explicitly exposes spectral and phase information essential for object \rev{localization}.
%
%
%
The complex-valued STFT coefficients are represented by their real and imaginary parts, 
yielding $\mathcal{A}_{\text{STFT}} = g_{\text{STFT}}(\mathcal{A}) \in \mathbb{R}^{M \times 2 \times T \times F}$,
where $T$ is the number of time frames in the STFT, and $F$ is the number of frequency bins.

\textbf{Ego-Velocity Encoding.}
\rev{In addition to acoustic features, 
the network is conditioned on
ego speed $v_{\text{ego}}$,
obtained from the vehicle’s proprioceptive velocity sensors (e.g., IMU). 
Relative motion between
the ego vehicle and surrounding sound sources can alter the temporal
and spectral characteristics of the received acoustic signals, 
including
Doppler shifts. 
Higher ego speeds also generally increase wind and
tire--road noise, 
which can mask sounds from surrounding vehicles
and reduce the effective signal-to-noise ratio. 
Conditioning on ego
speed provides context for adapting acoustic feature processing to
these motion and noise variations across driving speeds.}
Feature-wise linear modulation (FiLM) \cite{perez2018film} is applied to encode the velocity information \rev{recorded with each sample} $v_{\text{ego}}$ into $\mathcal{A}_{\text{STFT}}$, 
\begin{equation}
\tilde{\mathcal{A}}_{\text{STFT}} = \bigl(1+ \gamma( v_{\text{ego}}) \bigr) \odot \mathcal{A}_{\text{STFT}} + \beta(v_{\text{ego}}),
\label{eq:vel_ego}
\end{equation}
where $\gamma(\cdot)$ and $\beta(\cdot)$ are two learned functions for ego vehicle's velocity, 
and $\odot$ denotes element-wise multiplication with broadcasting.

\textbf{Backbone.}
\rev{A ResNet \cite{he2016resnet} serves} as backbone $g_{\text{backbone}(\cdot)}$ to extract deep hierarchical features from image-like STFT representations $\tilde{\mathcal{A}}_{\text{STFT}}$. 
%
%
Through successive convolutional and residual layers, the backbone learns to extract and integrate these DOA cues, 
yielding feature maps which encode the surrounding time–frequency–dependent directional information.
\rev{The resulting representation can therefore be interpreted} as a learned Time–Frequency DOA (TF-DOA) feature map, denoted by 
$\tilde{\mathcal{A}}_{\text{TF-DOA}} = g_{\text{backbone}}(\tilde{\mathcal{A}}_{\text{STFT}}) \in \mathbb{R} ^ {N_{\theta,\text{DOA}} \times T' \times F'}$, 
\rev{where $N_{\theta,\mathrm{DOA}}$ denotes the channel dimension
of the learned representation, and $T'<T$ and $F'<F$ denote its
temporal and frequency dimensions.}

\textbf{Head.}
After obtaining the $\tilde{\mathcal{A}}_{\text{TF-DOA}}$, 
\rev{a lightweight prediction head $g_{\text{head}}(\cdot)$ generates} a polar heatmap that jointly represents object center confidence over direction and range.
The head aggregates TF-DOA features 
and outputs a heatmap over discretized direction and range bins, where each cell represents the confidence of object center presence. 
\rev{Formally, the prediction is
$\hat{Y}=g_{\mathrm{head}}
(\tilde{\mathcal{A}}_{\mathrm{TF\text{-}DOA}})
\in[0,1]^{N_\theta\times N_r}$.}

\textbf{Loss Function.}
\label{subsec:loss}
\rev{The predicted polar heatmap is supervised using the mean squared
error (MSE) loss} between the network output and the ground-truth labels, namely,
\begin{equation}
\mathcal{L}_{\mathrm{MSE}} = \frac{1}{N_\theta N_r} \sum_{i=1}^{N_\theta} \sum_{j=1}^{N_r} (\hat{Y}_{ij} - Y_{ij})^2.
\label{eq:mse_loss}
\end{equation}

\rev{\textbf{Object Decoding via Heatmap Clustering.}}
\rev{Object centers are decoded from the predicted polar heatmap
$\hat{Y}$ by retaining cells with $\hat{Y}_{ij}>\tau_c$ and grouping
adjacent retained cells into $K$ connected components. Let
$\mathcal{D}_k$ denote the set of grid indices in the $k$-th component.
Each component yields one predicted object center, computed as the
confidence-weighted mean of its polar coordinates:}
\begin{equation}
\rev{
\hat{\mathbf{p}}_k =
\frac{\sum_{(i,j)\in\mathcal{D}_k}
\hat{Y}_{ij}\,(\theta_i,r_j)}
{\sum_{(i,j)\in\mathcal{D}_k}\hat{Y}_{ij}},
}
\end{equation}

\rev{where $\hat{\mathbf{p}}_k=(\hat{\theta}_k,\hat{r}_k)$ is the
predicted object center in polar coordinates. If no cells exceed
the threshold, $K=0$ and no objects are localized.}




\section{Experiment}
\label{sec:exp:acouticbev_obj_det}

\subsection{Experiment Setup}

\textbf{Evaluation Metrics.}
Predicted object centers are matched to ground-truth centers using a one-to-one assignment based on predefined angular and range thresholds, denoted by $\tau_\theta$ and $\tau_r$, in polar space. 
A prediction is considered a true positive (TP) if it is successfully matched to a ground-truth object within these thresholds, 
a false positive (FP) if it does not match any ground-truth object, 
and a false negative (FN) if a ground-truth object is not matched.
Based on these definitions, we evaluate localization performance using 
\rev{the F1-score, $F1^{\tau_\theta, \tau_r}$, defined as the harmonic
mean of precision and recall under the angular and range matching
thresholds $\tau_\theta$ and $\tau_r$.} 
\rev{$\tau_\theta = 5^\circ$ and $\tau_r = 5 \:\mathrm{m}$ are default thresholds}.
Furthermore, 
for true positive predictions, 
we report TP metrics, 
namely the mean Average Distance Error (mADE) and mean Average Angle Error (mAAE), 
which measure the average range and angular deviations between matched predictions and ground-truth object centers.

\textbf{Hyperparameters.}
In this work,
\rev{the acoustic perception field is defined in polar coordinates as}
$\Theta = [0, 360^\circ]$ and $R = [0, 30\,\mathrm{m}]$.
The polar grid is discretized with angular and range resolutions
$\Delta_\theta = 1^\circ$ and $\Delta_r = 0.5\,\mathrm{m}$,
resulting in $N_\theta = 360$ angle bins and $N_r = 60$ range bins.

For AcousticBEV, we use all microphone signals from the AcousticScenes \textit{train} set as input for training.
Each sample consists of recordings from $M=32$ microphones with an inherent sampling frequency $f=48$ kHz over a temporal window of $\Delta t = 100\,\mathrm{ms}$.
We follow the common setup of STFT \cite{he2021deep} and only focus on traffic-related frequencies in the range $[100,\:8000]$ Hz \cite{salamon2014dataset}, yielding $T = 19$ time frames and $F = 84$ frequency bins. 
\rev{For the range-dependent soft labels,
we set $\sigma_r^{\min}=1.5\,\mathrm{m}$,
$\sigma_r^{\max}=3\,\mathrm{m}$,
$\sigma_\theta^{\min}=4^\circ$, and
$\sigma_\theta^{\max}=6^\circ$.}
%
During object decoding, 
\rev{a confidence threshold of $\tau_c = 0.25$ is applied.}

\textbf{Training \& Inference.} 
AcousticBEV is trained end-to-end with the Adam optimizer with a learning rate of $3\times10^{-5}$ and a weight decay of $5\times10^{-4}$ for $100$ epochs.
Early stopping is applied if the validation performance does not improve for 10 consecutive epochs.
All experiments are conducted on a single NVIDIA RTX 3090 GPU. 
\rev{AcousticBEV contains 3.3M parameters, 
requires 2.309 GFLOPs per sample and achieves an average inference latency of 1.571 ms over 100 runs (\textgreater 500 FPS).}

\begin{table}[!t]
\centering
\caption{\rev{\textbf{Ablation of velocity-encoding (VE) strategies on the AcousticScenes \textit{test} split, by ego speed.}
Mean \textpm\ std over three training seeds. Group headers give the number of test scenes / samples.
(\textbf{Best}, \underline{second best}, \textit{default}.)}}
\label{tab:ablation_ve_test}
\footnotesize
\setlength{\tabcolsep}{2pt}
\renewcommand{\arraystretch}{1.05}
\begin{tabular*}{\columnwidth}{@{\extracolsep{\fill}}lccc@{}}
\toprule
Method & $F1^{5^\circ,5\mathrm{m}}$ (\%) $\uparrow$
       & mADE (m) $\downarrow$ & mAAE (\textdegree) $\downarrow$ \\
\midrule
\multicolumn{4}{@{}l@{}}{\textbf{Static}\quad \rev{13 scenes, 412 samples}} \\
w/o VE & \rev{47.4\std{2.3}}
       & \rev{\underline{1.91}\std{0.14}} & \rev{1.85\std{0.09}} \\
MLP    & \rev{\textbf{48.3}\std{3.7}}
       & \rev{2.03\std{0.15}} & \rev{1.94\std{0.08}} \\
Fourier & \rev{47.2\std{4.1}} & \rev{1.97\std{0.10}} & \rev{\underline{1.83}\std{0.07}} \\
\textit{FiLM} & \rev{\underline{47.5}\std{3.3}}
       & \rev{\textbf{1.81}\std{0.10}} & \rev{\textbf{1.82}\std{0.06}} \\
\midrule
\multicolumn{4}{@{}l@{}}{\textbf{10 km/h}\quad \rev{2 scenes, 65 samples}} \\
w/o VE & \rev{\underline{52.2}\std{4.0}}
       & \rev{\underline{2.11}\std{0.11}} & \rev{2.18\std{0.09}} \\
MLP    & \rev{47.6\std{2.3}}
       & \rev{2.20\std{0.22}} & \rev{1.85\std{0.23}} \\
Fourier & \rev{47.5\std{10.6}} & \rev{2.23\std{0.22}}
        & \rev{\underline{1.84}\std{0.16}} \\
\textit{FiLM} & \rev{\textbf{57.4}\std{1.3}}
       & \rev{\textbf{2.01}\std{0.30}} & \rev{\textbf{1.66}\std{0.27}} \\
\midrule
\multicolumn{4}{@{}l@{}}{\textbf{30 km/h}\quad \rev{5 scenes, 145 samples}} \\
w/o VE & \rev{39.4\std{3.2}}
       & \rev{1.95\std{0.17}} & \rev{1.96\std{0.19}} \\
MLP    & \rev{36.5\std{4.0}} & \rev{2.14\std{0.04}} & \rev{\underline{1.94}\std{0.13}} \\
Fourier & \rev{\underline{40.6}\std{5.7}}
        & \rev{\underline{1.91}\std{0.31}} & \rev{\textbf{1.90}\std{0.07}} \\
\textit{FiLM} & \rev{\textbf{43.1}\std{5.1}}
       & \rev{\textbf{1.67}\std{0.16}} & \rev{\underline{1.94}\std{0.11}} \\
\midrule
\multicolumn{4}{@{}l@{}}{\textbf{50 km/h}\quad \rev{4 scenes, 110 samples}} \\
w/o VE & \rev{\underline{35.8}\std{2.8}}
       & \rev{\underline{1.90}\std{0.27}} & \rev{\textbf{1.94}\std{0.10}} \\
MLP    & \rev{29.6\std{2.5}} & \rev{2.19\std{0.22}} & \rev{2.19\std{0.07}} \\
Fourier & \rev{30.5\std{4.7}}
        & \rev{\underline{1.90}\std{0.14}} & \rev{2.11\std{0.20}} \\
\textit{FiLM} & \rev{\textbf{36.7}\std{2.9}}
       & \rev{\textbf{1.71}\std{0.15}} & \rev{\underline{2.09}\std{0.25}} \\
\midrule
\multicolumn{4}{@{}l@{}}{\textbf{Total} \quad  \rev{24 scenes, 732 samples}} \\
w/o VE & \rev{\underline{44.7}\std{2.0}}
       & \rev{\underline{1.93}\std{0.05}} & \rev{1.92\std{0.07}} \\
MLP    & \rev{43.2\std{3.1}} & \rev{2.08\std{0.12}} & \rev{1.95\std{0.03}} \\
Fourier & \rev{43.5\std{4.6}}
        & \rev{1.98\std{0.12}} & \rev{\underline{1.87}\std{0.02}} \\
\textit{FiLM} & \rev{\textbf{46.0}\std{1.0}}
       & \rev{\textbf{1.80}\std{0.03}} & \rev{\textbf{1.85}\std{0.06}} \\
\bottomrule
\end{tabular*}
\end{table}

\subsection{Baselines}

\rev{Beamforming and
MUSIC~\cite{schmidt1986multiple} serve as classic baselines for direction prediction only. 
For metric
BEV object localization, we also evaluate two learning-based baselines:
\mbox{M-BF+Head} and
SELDNet~\cite{adavanne2018seldnet,shimada2022multi}.}


\textbf{Beamforming (BF)}~\cite{sarradj2012three}
\rev{estimates acoustic energy over a predefined angular grid, with
high-response regions indicating candidate source directions.
These responses are mapped onto a circle of fixed radius
$20\,\mathrm{m}$ in the BEV plane.
This fixed-radius representation provides directional information
without estimating source range.}


\textbf{MUSIC}~\cite{schmidt1986multiple}
\rev{is a subspace-based DOA estimation method that computes a
spatial pseudo-spectrum over a predefined angular grid.
As with BF, 
its responses are mapped onto a circle of fixed radius
$20\,\mathrm{m}$ in the BEV plane. This representation
encodes source direction without estimating range.}


\textbf{Multi-Radius Beamforming with the Detection Head (M-BF+Head)}
\rev{extends BF to metric BEV object localization. We stack
beamforming responses computed at multiple predefined radii and
feed them into a detection head identical to that used in AcousticBEV.
M-BF+Head is trained with the same protocol as AcousticBEV for a controlled comparison
between beamforming and learned TF-DOA features.}


\textbf{SELDNet}~\cite{adavanne2018seldnet}
\rev{formulates localization as direct regression using a CRNN with
CNN and GRU layers. MIC-GCC features and a
Multi-ACCDOA-style head~\cite{shimada2022multi} are used, adapted to regress
object direction and range. This yields sparse localization
estimates, whereas AcousticBEV regresses a dense polar heatmap
from which object centers are decoded.}
%
%

\begin{figure}[b]
    \centering
    \includegraphics[width=\columnwidth]{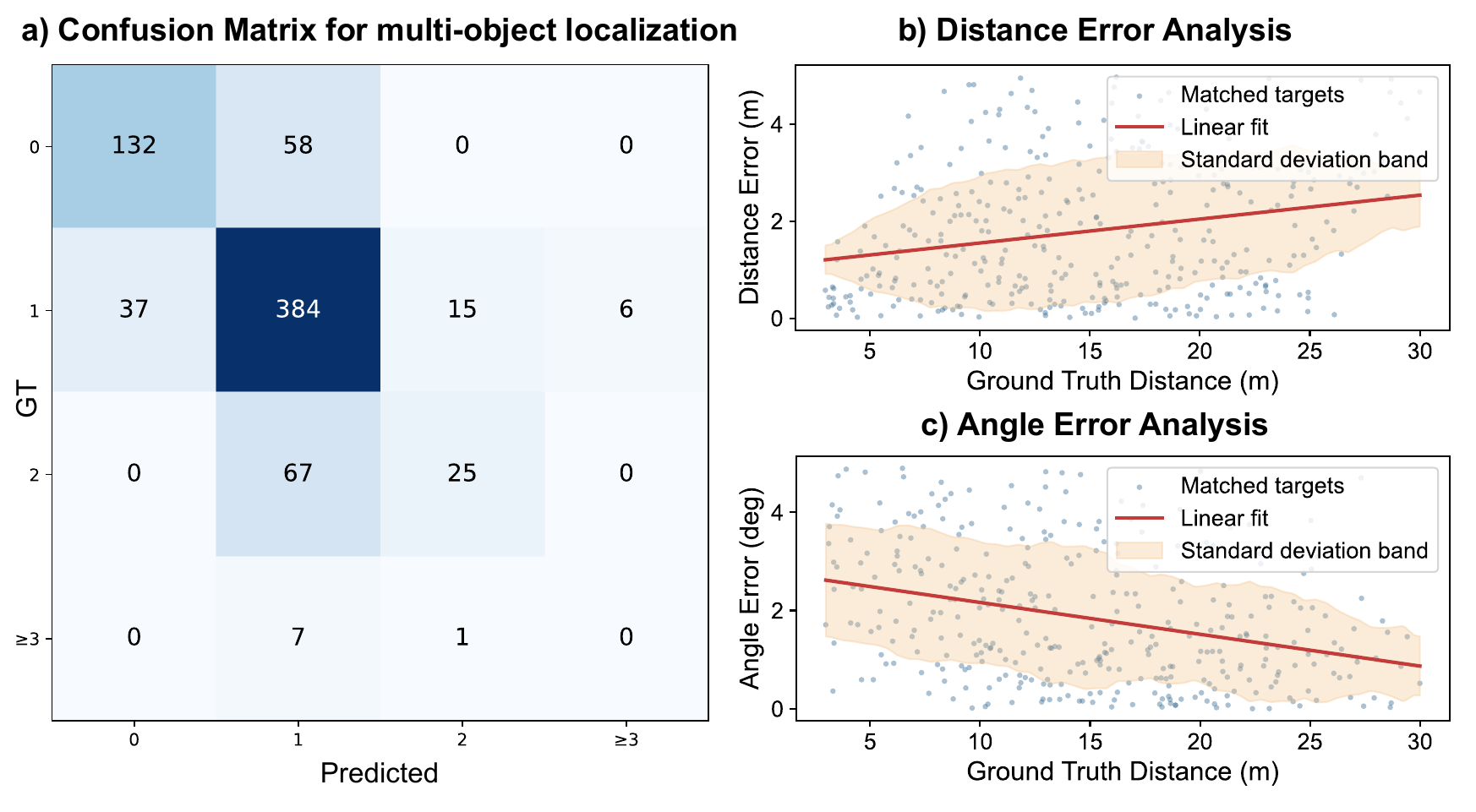}
    \caption{\textbf{Localization Analysis on AcousticScenes \textit{test} split.}}
    \label{fig:error_analysis}
\end{figure} 

\subsection{Acoustic BEV Object Localization}
We demonstrate the acoustic BEV object localization performance of AcousticBEV and other baselines on the AcousticScenes \textit{test} set in Tab.~\ref{tab:compare_baseline}. 
\rev{Unless otherwise stated, results of learning-based methods are reported as mean\std{standard deviation} over three independent training runs, 
and all analyses are based on these results. BF and MUSIC are deterministic and reported as single values.}
%
For \textit{angle estimation}, BF achieves $F1^{5^\circ}$ scores of 53.8\% in the `Easy' category and 37.3\% in the `Difficult' category, respectively.
%
AcousticBEV achieves $F1^{5^\circ}$ scores of \rev{56.3\std{1.6}\%} and \rev{52.1\std{1.5}\%} in the `Easy' and `Difficult' cases, respectively, \rev{outperforming BF and MUSIC in both scenarios.}
\rev{In contrast, for direction-only estimation, the learning-based baselines M-BF+Head and SELDNet underperform classic BF.}
%
\rev{For \textit{acoustic BEV object localization}, AcousticBEV
achieves $F1^{5^\circ,5\,\mathrm{m}}$ scores of 47.7\std{1.2}\% and
45.1\std{1.4}\% in `Easy' and `Difficult', respectively,
outperforming M-BF+Head (30.1\std{1.1}\% and 21.6\std{3.2}\%) and
SELDNet~\cite{adavanne2018seldnet} (33.1\std{5.7}\% and 25.5\std{2.6}\%).}
Regarding the TP metrics, 
AcousticBEV further achieves mADE with \rev{1.76\std{0.13}~$\mathrm{m}$} in `Easy' and \rev{1.81\std{0.06}$~\mathrm{m}$} in `Difficult' and mAAE with \rev{1.79\std{0.09}$^\circ$} in `Easy' and \rev{1.85\std{0.05}$^\circ$} in `Difficult'. All outperform the M-BF+Head and SELDNet.

\rev{Fig.~\ref{fig:error_analysis} a) presents AcousticBEV's
object-count confusion matrix for Easy and Difficult samples combined. The model correctly estimates the
count for most single-object samples, but sometimes predicts one
object in empty samples or only one when two are present.
The latter may reflect difficulty in resolving weak or overlapping
acoustic sources.}


\begin{figure}[t]
    \centering
    \includegraphics[width=0.9\columnwidth]{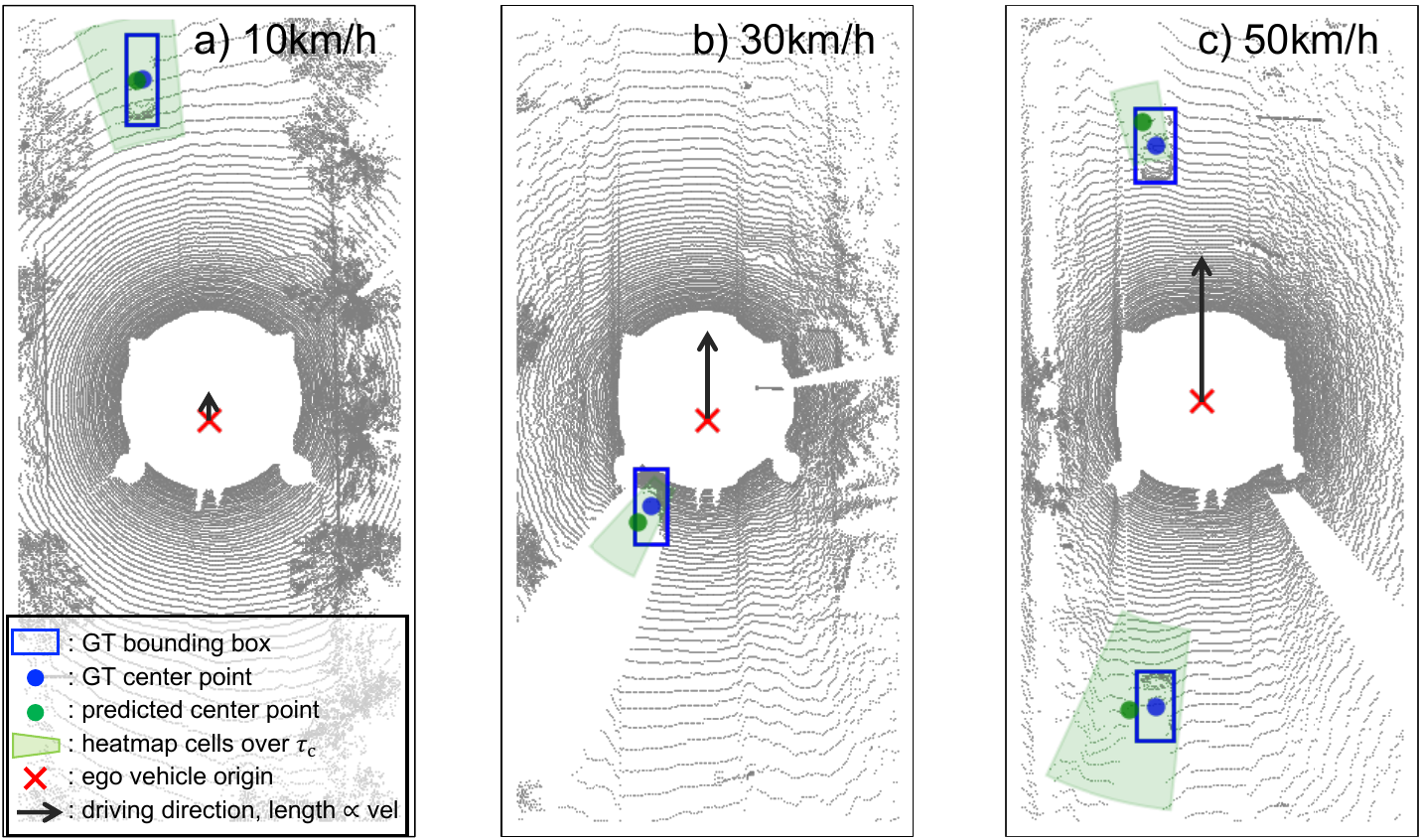}
    \caption{\textbf{Qualitative results of Acoustic BEV Object Localization under different velocities.} 
    The point clouds are shown for visualization purposes only.}
    \label{fig:moving_test}

\end{figure}

\begin{figure}[b]
    \centering
    \includegraphics[width=0.9\columnwidth]{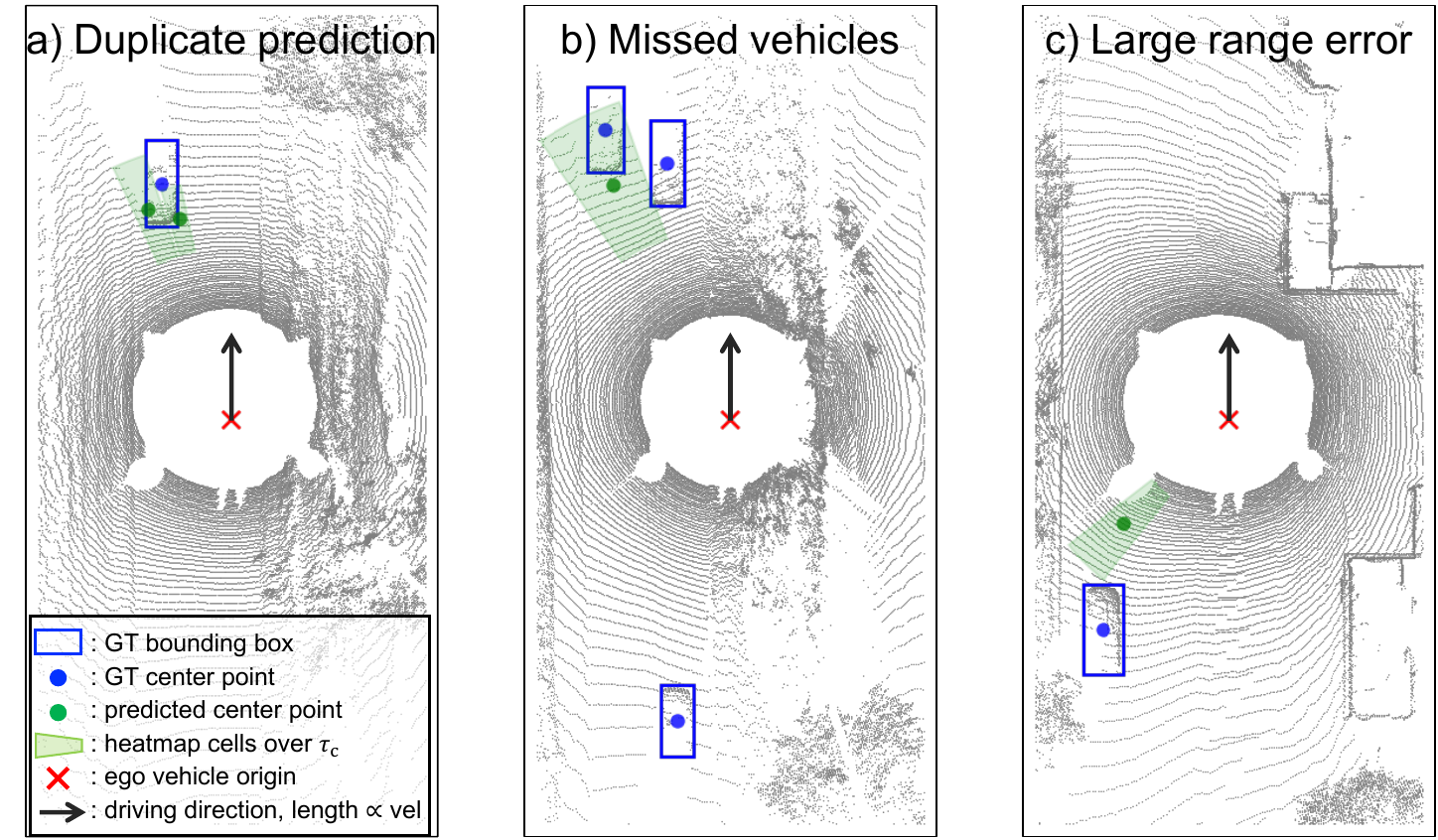}
    \caption{\rev{\textbf{Failure cases of AcousticBEV on the AcousticScenes \textit{test} split.}
a) Duplicate prediction for one vehicle. b) Missed vehicles in a multi-object scene. c) Range underestimation. Point clouds are shown for visualization only.}}
    \label{fig:failure_case}

\end{figure}

\subsection{Ablation \& Analysis}

\textbf{Acoustic BEV Object Localization at Different Velocities.}
In Tab.~\ref{tab:ablation_ve_test}, 
we further examine the acoustic BEV object localization performance of AcousticBEV (with default FiLM velocity encoding module) under different ego vehicle velocities on AcousticScenes \textit{test} set:
$0$ km/h (static), 
$10$ km/h, 
$30$ km/h, 
and $50$ km/h. 
%
%
%
\rev{Performance decreases at $30\,\mathrm{km/h}$ and $50\,\mathrm{km/h}$,
potentially reflecting changes in acoustic observations and
noise conditions at higher speeds. FiLM-based velocity
encoding achieves higher mean $F1$ than the variant without VE
across all evaluated ego velocities, suggesting the value of
ego-velocity context for acoustic localization.}

\begin{table}[t]
\centering
\caption{\rev{\textbf{Scene-bootstrap evaluation on the
AcousticScenes test split.}
Values are point estimates on the full test split,
with 95\% scene-bootstrap CIs in brackets. Bottom rows report paired
differences (AcousticBEV $-$ baseline);
$F1$ differences are in percentage points.}}
\label{tab:scene_bootstrap}
\footnotesize
\renewcommand{\arraystretch}{1.1}
\setlength{\tabcolsep}{1pt}

\newcommand{\bootci}[2]{#1{\scriptsize\hspace{0.5pt}[#2]}}

\begin{tabular*}{\columnwidth}{@{\extracolsep{\fill}}lccc@{}}
\toprule
\rev{\textbf{Method}} &
\rev{$F1^{5^\circ,5\mathrm{m}}$ (\%) $\uparrow$} &
\rev{mADE (m) $\downarrow$} &
\rev{mAAE ($^\circ$) $\downarrow$} \\
\midrule

\rev{M-BF+Head}
& \rev{\bootci{24.3}{18.8,30.3}}
& \rev{\bootci{2.35}{2.11,2.57}}
& \rev{\bootci{2.08}{1.81,2.32}} \\

\rev{SELDNet~\cite{adavanne2018seldnet,shimada2022multi}}
& \rev{\bootci{28.1}{22.6,33.8}}
& \rev{\bootci{2.21}{2.07,2.37}}
& \rev{\bootci{2.32}{2.16,2.49}} \\

\rev{\textbf{AcousticBEV} (ours)}
& \rev{\bootci{\textbf{46.0}}{41.9,50.2}}
& \rev{\bootci{\textbf{1.80}}{1.68,1.91}}
& \rev{\bootci{\textbf{1.85}}{1.68,2.01}} \\

\midrule

\rev{$\Delta$ vs M-BF+Head}
& \rev{\bootci{+21.7}{15.8,27.5}}
& \rev{\bootci{-0.55}{-0.78,-0.33}}
& \rev{\bootci{-0.23}{-0.54,0.12}} \\

\rev{$\Delta$ vs SELDNet}
& \rev{\bootci{+17.9}{12.6,22.9}}
& \rev{\bootci{-0.41}{-0.58,-0.27}}
& \rev{\bootci{-0.47}{-0.70,-0.25}} \\

\bottomrule
\end{tabular*}
\end{table}

\textbf{Ablation of Ego-Velocity Encoding.}
%
Tab.~\ref{tab:ablation_ve_test} compares four ego-velocity encoding (VE) strategies: no velocity encoding (w/o VE), Fourier, MLP, and FiLM. 
Fourier VE encodes the ego velocity using fixed sinusoidal Fourier features, whereas MLP VE learns a velocity embedding through an MLP.
%
%
%
FiLM shows consistently competitive $F1$ scores and TP metrics across different velocities.
\rev{\textbf{Ablation of SMA Design.}}
\rev{Two 16-microphone configurations have been tested,
each obtained by
symmetrically subsampling the SMA: one spans both layers,
while the other uses a single layer. The two-layer configuration
achieves an $F1^{5^\circ,5\,\mathrm{m}}$ of $41.7\%$, compared to
$38.6\%$ for the single-layer configuration, supporting the benefit
of the two-layer design. These results are based on a single run.}

\textbf{Angle and Distance Error Analysis.} 
Fig.~\ref{fig:error_analysis} b) and c) show the distribution of distance and angle errors \rev{for} different ground-truth distances.
As shown in the upper figure, as the ground-truth object distance increases, the distance error increases approximately linearly, 
while the lower figure shows a roughly linear decreasing trend for the angle error.
These observations align with our initial hypothesis of the angle and distance uncertainty $\sigma_{\theta}$ and $\sigma_r$ in Eqs. \eqref{eq:signma_r} and \eqref{eq:signma_theta}.

\rev{\textbf{Qualitative Analysis.}}
\rev{Fig.~\ref{fig:moving_test} shows qualitative examples of AcousticBEV at different ego speeds, and Fig.~\ref{fig:failure_case} illustrates typical failure modes.}

\begin{figure}[b]
    \centering
    \includegraphics[width=\columnwidth]{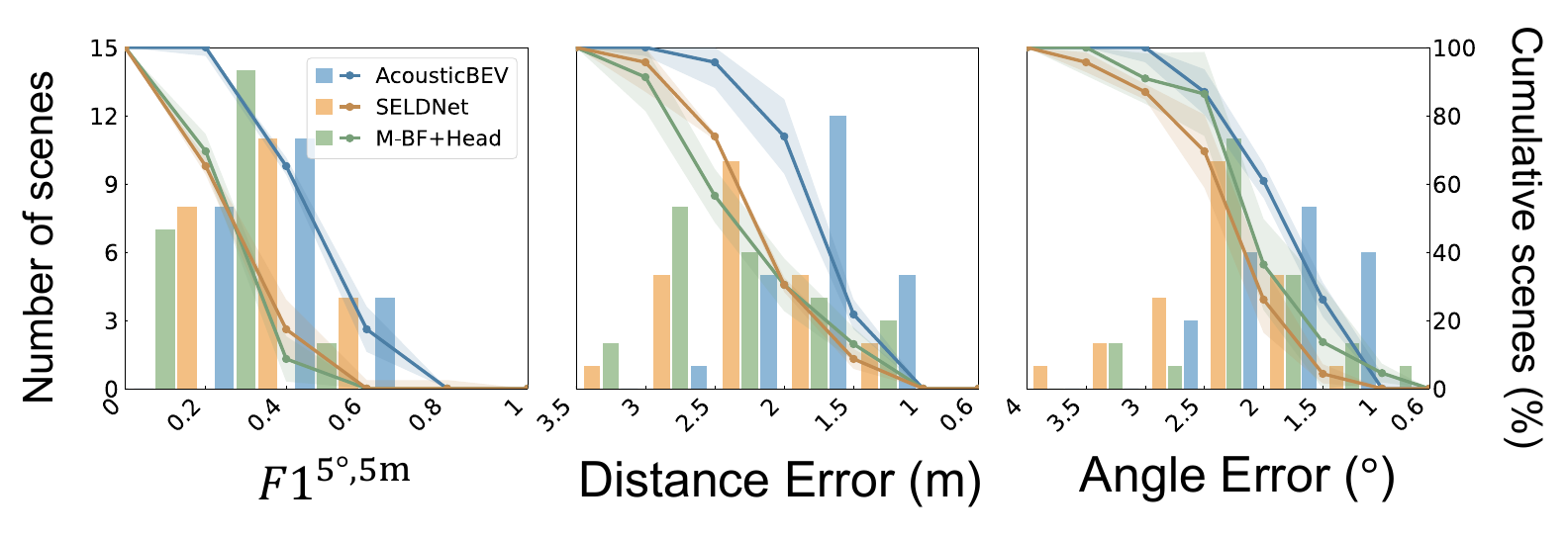}

\caption{\rev{\textbf{Scene-level performance on the AcousticScenes
\emph{test} split.} Bars show scene counts; cumulative curves show the
mean percentage of scenes meeting each performance threshold,
with shaded bands indicating $\pm1$ standard deviation across
three independent training runs. Rightward values indicate
better performance.}}
\label{fig:scene_evaluation}
\end{figure}

\subsection{\texorpdfstring{\rev{Scene-level Evaluation}}{Scene-level Evaluation}}
\rev{Finally, the statistical uncertainty is quantified using scene bootstrapping, and performance variations across scenes are examined.}

\rev{\textbf{Scene Bootstrap Analysis.}}
%
\rev{To account for within-scene temporal dependence,
the 24 test scenes are resampled with replacement for 10,000 bootstrap replicates, 
using identical scene draws across methods. 
For each replicate, one checkpoint per method is selected independently, 
and pooled $F1$, mADE, mAAE, and paired differences (AcousticBEV $-$ baseline) are recomputed. 
We report 95\% bootstrap uncertainty intervals given by the 2.5th and 97.5th percentiles.}


%

\rev{Tab.~\ref{tab:scene_bootstrap} shows $F1$ gains of 21.7 and
17.9 percentage points over M-BF+Head and SELDNet, respectively.
The paired bootstrap intervals favor AcousticBEV in $F1$ and
mADE against both baselines, and in mAAE against SELDNet.
The mAAE difference against M-BF+Head has an interval spanning
zero. These results support AcousticBEV's gains in localization
and range estimation.}




\rev{\textbf{Per-Scene Performance.}
Fig.~\ref{fig:scene_evaluation} complements these aggregate results
with per-scene metric distributions and cumulative curves. 
The overall rightward shift of
AcousticBEV's $F1$ and error curves indicates higher scene-level $F1$ compared to its baselines. 
}

\section{Conclusions}
\label{sec:conclusion}

We introduced the \textit{AcousticScenes} dataset, 
which contains \rev{$4{,}980$} synchronized microphone recordings, point clouds, and 3D bounding box annotations,
providing a benchmark for studying acoustic perception in outdoor environments.
We also presented a BEV object \rev{localization} method, \textit{AcousticBEV}, to localize nearby road users in challenging outdoor settings \rev{using a vehicle-mounted Surround Microphone Array as the sole exteroceptive sensor}.
Our experimental results demonstrate that AcousticBEV outperforms \rev{both} classical signal-processing and \rev{learning-based} baselines, 
achieving $F1^{5^\circ,5\,\mathrm{m}}$ scores of \rev{47.7\std{1.2}\%}  on `Easy' scenarios and \rev{45.1\std{1.4}\%} on `Difficult' cases.
Our study has empirically demonstrated that a low-cost vehicle-mounted microphone array can meaningfully contribute to spatial localization in intelligent vehicles, not only when the ego vehicle is stationary but also while driving, highlighting the potential of microphones as a complementary sensing modality for future autonomous systems.

\textbf{Limitations and Future Work.}
\rev{AcousticScenes lacks highly congested traffic and complex road
layouts, such as narrow T-junctions. Future directions include
broader evaluation across driving tasks and fusion of acoustic cues
with camera and LiDAR data under adverse weather conditions.}

\bibliographystyle{IEEEtran}
\bibliography{main}


\end{document}